\documentclass[
]{ceurart}

\usepackage{listings}
\usepackage{graphicx} 
\usepackage{geometry}
\usepackage{amsmath}
\usepackage{tikz}
\usepackage{amsmath}
\usepackage{algpseudocode}
\usepackage{algorithm}
\usepackage{subcaption}
\usepackage{multicol}
\usepackage{setspace}
\usepackage{todonotes}
\usepackage{pdfpages}
\usepackage{pdflscape}
\usepackage{rotating}
\usepackage{amsfonts}
\usepackage{tabularx} 
\newcommand{\localdata}{\mathcal{D}_i}
\newcommand{\instance}{\mathbf{x}_i}
\newcommand{\explanation}{g_i}
\newcommand{\dissimilarity}{\Delta_{\mathcal{X}}}
\begin{document}

\copyrightyear{2024}
\copyrightclause{Copyright for this paper by its authors.
  Use permitted under Creative Commons License Attribution 4.0
  International (CC BY 4.0).}

\conference{HI-AI@KDD, Human-Interpretable AI Workshop at the KDD 2024, 26$^{th}$ of August 2024, Barcelona, Spain }

\title{MASALA: Model-Agnostic Surrogate Explanations by Locality Adaptation}


%
\author[1]{Saif Anwar}[%
orcid=0009-0000-5763-4416,
email=saif.anwar@warwick.ac.uk,
]
\cormark[1]
\address[1]{Department of Computer Science, University of Warwick, United Kingdom}
\address[2]{School of Engineering, University of Warwick, United Kingdom}

\author[1]{Nathan Griffiths}[%
orcid=0000-0002-6406-8632,
]

\author[1]{Abhir Bhalerao}[%
orcid=0000-0001-8830-329X,
]

\author[2]{Thomas Popham}[%
orcid=0000-0002-4567-9534,
]


\begin{abstract}
Existing local Explainable AI (XAI) methods, such as LIME, select a region of the input space in the vicinity of a given input instance, for which they approximate the behaviour of a model using a simpler and more interpretable surrogate model. The size of this region is often controlled by a user-defined locality hyperparameter. In this paper, we demonstrate the difficulties associated with defining a suitable locality size to capture impactful model behaviour, as well as the inadequacy of using a single locality size to explain all predictions. We propose a novel method, MASALA, for generating explanations, which automatically determines the appropriate local region of impactful model behaviour for each individual instance being explained. MASALA approximates the local behaviour used by a complex model to make a prediction by fitting a linear surrogate model to a set of points which experience similar model behaviour. These points are found by clustering the input space into regions of linear behavioural trends exhibited by the model. We compare the \emph{fidelity} and \emph{consistency} of explanations generated by our method with existing local XAI methods, namely LIME and CHILLI. Experiments on the PHM08 and MIDAS datasets show that our method produces more faithful and consistent explanations than existing methods, without the need to define any sensitive locality hyperparameters.
\end{abstract}

\begin{keywords}
Exlainable AI (XAI), Interpretable Machine Learning, Explanation, Model-Agnostic, Post-Hoc, Local Linear Modelling
\end{keywords}

\maketitle

\section{Introduction}
Many Machine Learning (ML) methods are treated as \emph{black-boxes} because of their complex and often incomprehensible behaviour. As a result, there is tentative adoption in high-risk domains, such as healthcare, finance, and defence. There is a demand from stakeholders to establish trust in a model, since an incorrect decision may have serious consequences \cite{devittMethodEthicalAI2021,singhCurrentChallengesBarriers2020, preeceStakeholdersExplainableAI2018}. Explainable AI (XAI) methods aim to provide explanations for the predictions produced by a model and make transparent its behaviours \cite{dosilovicExplainableArtificialIntelligence2018a}. In this paper, we define an \emph{explanation} to be an interpretable representation of a \emph{base model}'s decision-making process, such as in the form of feature importance scores or a set of decision rules. 

Generally, XAI techniques can be divided into inherently-interpretable models and post-hoc methods. The former involves developing model architectures which are interpretable by design and do not require extensive effort to understand the reasoning for a given output \cite{tangExplainableSpatioTemporalGraph2023,goerigkFrameworkInherentlyInterpretable2023}. However, it is generally agreed that limiting complexity, for the sake of interpretability, may hinder performance when compared to more complex black-box models \cite{Loyola-Gonzalez_2019, Wanner_Herm_Heinrich_Janiesch_2022}. 

Post-hoc XAI methods generate explanations for pre-trained black-box models, often in a model-agnostic manner \cite{molnar2022}. Popular methods typically explain predictions through feature importance scores, either on a global or local scale \cite{ribeiroWhyShouldTrust2016,lundbergUnifiedApproachInterpreting2017,friedmanElementsStatisticalLearning}. Global methods explain general model behaviour for all datapoints, whereas local methods explain the model behaviour used to make a specific prediction. A local explanation is constrained to a \emph{locality}, namely a region of the input space surrounding the instance for which the prediction is being explained, which we call the \emph{target instance}.
\begin{figure*}[!t]
  \centering
  \includegraphics[width=14.5cm]{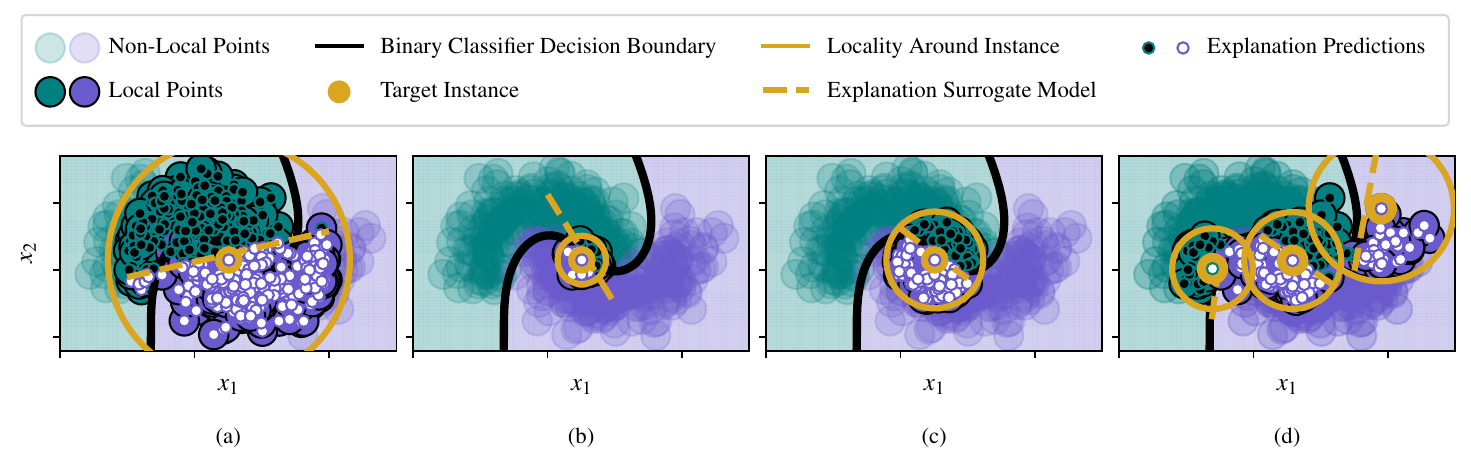} 
  \caption[]{\label{fig:varying_localities}a) A surrogate fit within a locality which is too large leading to an inaccurate linear approximation of the non-linear decision boundary b) A surrogate fit within a locality which is too small and therefore, captures irregularities rather than the impactful linear trend. c) An explanation generated within an appropriate sized locality, which represents the true model behaviour in the immediate vicinity of the target instance. d) Appropriate explanations generated for three instances using localities of different sizes}
  \end{figure*}
Some local XAI methods fit an inherently interpretable surrogate model in a locality around a target instance, where the interpretation of the surrogate model is the explanation, such as the coefficients of a regression model or rules established in a decision tree.

LIME \cite{ribeiroWhyShouldTrust2016} is a popular approach that fits a surrogate model to perturbations of a target instance that are generated by randomly sampling a Gaussian kernel centered around the target instance. The width of the kernel, which is manually defined by the user and is fixed for all explanations, controls the expanse of the perturbations and therefore the size of the explanation locality. It has been shown that a surrogate model fit to perturbations sampled from an inappropriately sized Gaussian kernel may not be representative of the base models training data and therefore does not represent the true base model behaviour \cite{dieberWhyModelWhy2020}. CHILLI \cite{anwarCHILLIDataContextaware2023} is an adaptation of LIME which addresses some of these issues. Perturbations are generated according to the distribution of the model training data in the vicinity of the target instance. However, CHILLI also requires the user to define the locality of each explanation. The locality size must be defined in such a way that it is not too large, where the explanation ignores model intricacies, and not too small, where the explanation focuses on anomalous fluctuations rather than more dominant behavioural trends. This is illustrated in Figure \ref{fig:varying_localities}, which shows how the locality size affects the behaviour of the surrogate model and how a fixed locality may not be appropriate for all instances. Since LIME and CHILLI use non-deterministic perturbation generation methods, explanations for the same prediction may differ, leading to a lack of consistency that undermines the trustworthiness of the explanation method \cite{zhaoBayLIMEBayesianLocal2021, tanGLIMEGeneralStable2023, laugelDefiningLocalitySurrogates2018}.

In this paper, we propose \textbf{M}odel-\textbf{A}gnostic \textbf{S}urrogate expl\textbf{A}nations by \textbf{L}ocality \textbf{A}daptation (MASALA), a novel post-hoc XAI method that automatically finds the impactful local model behaviour surrounding a target instance. MASALA fits a Multivariate Linear Regression (MLR) surrogate model to a set of points that experience the same linear behaviour as the target instance, which it obtains by automatically detecting the linear regions of model behaviour across the input domain. The coefficients of the MLR represent feature relationships towards the target distribution. Since MASALA generates explanations using a deterministic clustering, explanations for the same instance are identical and therefore are consistent. Using the PHM08 and MIDAS datasets, we qualitatively and quantitatively demonstrate the ability of MASALA to generate explanations with higher fidelity and consistency than those produced by LIME and CHILLI. Our source code and data are available through the following repository: \url{https://github.com/saiffanwar/MASALA}

\section{Related Work}
Existing works have attempted to address locality issues by clustering similar points to fit linear surrogate models. Zafar et al.~\cite{zafarDLIMEDeterministicLocal2019} propose DLIME, which uses agglomerative Hierarchical Clustering to divide the training data into groups of similar points according to their Euclidean distance across all features. Points that are clustered together may not experience the same model behaviour since they may be close in some feature dimensions, but distant in others and therefore may not experience similar model behaviour across all features.

It has been shown that LIME generates inconsistent explanations when using a locality size that is too small, since explanations may focus on irregularities introduced by the randomly sampled perturbations. Gaudel et al.~propose s-LIME \cite{gaudelSLIMEReconcilingLocality2022}, which generates perturbations whose distance is proportional to the magnitude of the selected kernel width. However, this still requires the locality to be manually defined. Local Surrogate \cite{laugelDefiningLocalitySurrogates2018} avoids manual locality definition by generating perturbations around the decision boundary closest to the instance being explained, therefore approximating the model behaviour that led to the prediction. However, this may not be applicable to a regression problem where there is no decision boundary. In ALIME \cite{shankaranarayanaALIMEAutoencoderBased2019}, an autoencoder is trained as a weighting function used to decide whether perturbations are local to the target instance. Although this leads to more consistent explanations, the threshold for discarding points must be manually defined and is effectively equivalent to the kernel width hyperparameter in LIME.

\begin{figure*}[!t]
  \centering
  \includegraphics[width=12cm]{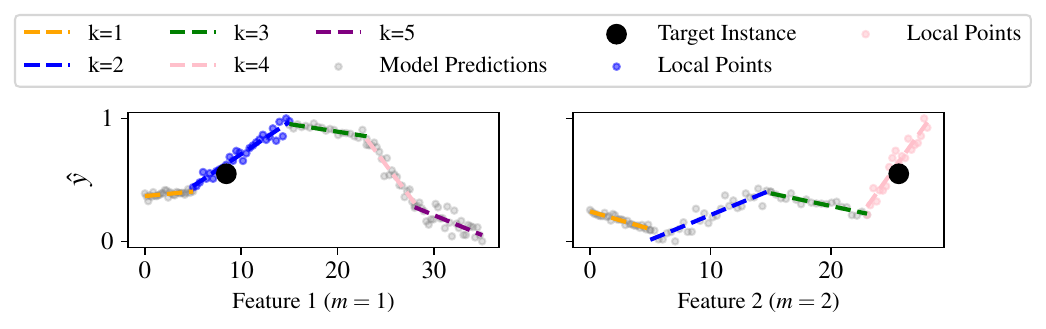}
  \caption{\label{fig:sample_clustering} Distribution of 2 features from the same dataset against the model predictions and clustered into linear regions.}
  \end{figure*}
\section{Methodology}


The goal of MASALA is to fit a linear surrogate model to a set of points that experience similar model behaviour to a specified target instance. We first formalise the problem and then describe the details of the method.
\subsection{Problem Definition}
\label{sec:problem_definition}
Consider a black-box base model $f$, which maps an $M$-dimensional input vector $\mathbf{x} \in \mathcal{X}$, to a scalar output $y \in \mathcal{Y}$, and is trained on a dataset $\mathcal{D}$. The prediction for a given target instance $\instance$ is explained by training a MLR surrogate model $\explanation$ on a subset of the training data $\localdata \in \mathcal{D}$, with target values being the predictions on $\localdata$ from the base model $f$. The linear coefficients of the MLR directly indicate the contribution of each feature towards the prediction. The selection of instances to include in $\localdata$ defines the locality of the explanation, since only the model behaviour used to make predictions for those instances will be approximated. Only instances that have similar feature values and experience similar model behaviour to the target instance should be included in $\localdata$. However, identifying a set $\localdata$ for a given $\instance$ is non-trivial, since the model behaviour may vary across the feature space. MASALA identifies an appropriate subset $\localdata$ for training a surrogate MLR model, by finding all instances that share the same region of linearity in the distribution of model predictions as the target instance $\instance$.
Since we assume that the base model behaviour is locally linear for some region around all data instances, we propose exhaustively clustering the distribution of each input feature against the model predictions into regions of linearity. The set of $K$ identified linear regions, or clusters, for a given feature dimension $m$, is denoted as $\mathcal{C}^m$, such that $\mathcal{C}^m = \{c_{k}^{m} | \forall {k \in K}\}$. Once a clustering has been obtained, the clusters within which a target instance $\instance$ falls, in each feature dimension, can be identified. This is denoted as $c^{m}_{k} (\instance)$ which indicates that in the distribution of feature $m$, the target instance falls within linear region $k$. The subset of instances used to train the surrogate $\explanation$ is then defined as the set of instances which share the same linear region in each feature dimension as the instance $\instance$, as formulated in Equation \ref{eq:localdata}. 
\begin{equation}
\localdata = \{ \mathbf{x}_j \in \mathcal{X} \vert c^{m}_{k}(\mathbf{x}_j) = c^{m}_{k}(\instance), \forall  m \in M \}
\label{eq:localdata}
\end{equation}
Figure \ref{fig:sample_clustering} shows the distributions of 2 features from the same dataset, against the respective model predictions, along with a target instance to be explained. Each feature has been clustered into a different number of linear regions with the target instance falling within the blue and pink regions for features $m=1$ and $m=2$ respectively. The set $\localdata$ contains the points that also fall within both the blue and pink clusters. Since $\localdata$ does not change for a given instance, explanations generated for the same instance are identical, thus preserving consistency.
\subsection{Local Linear K-Medoids Clustering}
\label{sec:MASALA}
We now present our method for identifying regions of linear model behaviour in the distribution of an input feature and the predictions from the base model. We consider each feature dimension individually, and will therefore omit the superscript $m$ in the remainder of this section. For example, $\mathcal{X}^m$ will be denoted as $\mathcal{X}$ and $c_{k}^{m}$ will be denoted as $c_k$.
\subsubsection{Pairwise Dissimilarity}
We apply an adapted K-medoids algorithm \cite{parkSimpleFastAlgorithm2009b} which  clusters datapoints based on their pairwise Euclidean distance. However, points that are close in the feature space, do not necessarily experience the same linear model behaviour. We introduce a custom distance measure in the form of pairwise dissimilarity $\dissimilarity$, which compares the local linearity, feature value, and local density of points. For each datapoint, $x_i$, a weighted Local Linear Regression (LLR) is performed on all points within its neighbourhood, $N(x_i)$, which is defined using a distance threshold in the feature space. Although this may be seen as defining a similar threshold to the kernel width locality parameter in LIME, the final explanation is much more robust to changes in the threshold since the weighted LLR automatically considers closer points with greater importance. The dissimilarity between two points $x_i$ and $x_j$ is calculated as 
\begin{equation}
    \dissimilarity(i,j) = {|| \mathbf{w}_i-\mathbf{w}_j||}_2 + D(x_i,x_j) + ||N(x_i)| - |N(x_j)||,
\end{equation}
\noindent
where $\mathbf{w}_i$ represents the vector of LLR model parameters for point $x_i$, and $|N(x_i)|$ denotes the number of points in the neighbourhood of $x_i$. The second term $D(i,j)$ is the difference in feature values of $x_i$ and $x_j$, which may be a custom measure dependent on the type of feature. Including this distance ensures that points with similar neighbourhood trends that are in different regions of the feature space, are not clustered together. The final term allows for the inclusion of local data density since two points may share a similar linear neighbourhood trend, but one neighbourhood may exhibit more sparsity than the other. A linear model fit to a sparse neighbourhood may be skewed by anomalous datapoints, and therefore should be considered with more caution. It should be noted that all terms are normalised in the range [0,1] such that they have equal contribution towards the dissimilarity measure. Points with the smallest values in $\dissimilarity$ will be close together in the feature space, have similar LLR model parameters, and have similar neighbourhood density.

Points are clustered according to their pairwise dissimilarity using the K-medoids algorithm \cite{parkSimpleFastAlgorithm2009b}. To allow for a deterministic clustering, medoids are initialised by evenly distributing K medoids across the sorted values from the data. A Linear Regression (LR) model is fit to the points within each cluster, where $a_k$ and $b_k$ are the LR parameters within cluster $k$. The cost, $J$, for a clustering $C$ is calculated according to Equation \ref{eq:clustering_cost}.
\begin{equation}
\label{eq:clustering_cost}
    J(C) = \sum_{k=1}^K \text{RMSE}(\{ a_k \cdot x + b_k , f(x) | \forall x \in C_k \})
\end{equation}
Thus, the cost is the sum of the RMSEs between predictions from the LR model and base model predictions within each cluster. Rather than randomly assigning a new medoid for a cluster, the clustering cost is calculated when each point is the medoid. The medoid which gives the lowest clustering cost is selected. This is repeated for all clusters with the algorithm halting when the clustering cost no longer changes after optimising all clusters and is also a deterministic process. A lower clustering cost reflects an ensemble of linear proxy models that is more faithful to the behaviour of the base model. 




\subsubsection{Automatically defining $K$}
It may seem that increasing the number of clusters $K$, would generate a more faithful ensemble of linear models. However, doing so leads to clusters with smaller coverage which may be overfit to erroneous behaviour, rather than to more impactful general linear trends. This effect can be worsened in sparse regions of the input space. The relationship between an input feature and model predictions will vary across features and datasets, and therefore the appropriate number of linear regions to cluster also varies. We propose an algorithm that automatically finds a suitable value for $K$, given a set of constraints, such that clusters must not overlap in the feature space, must contain a sufficient number of datapoints, and must cover a suitable range of values in the feature space. 

If a cluster is wholly contained within another cluster, the larger cluster will adopt all datapoints of the smaller cluster. If two clusters overlap each other, two new clusters replace them by dividing the points at the midpoint of the overlapping clusters' intersection. Once all clusters have been separated, they each occupy a unique range of values in the feature space.

To avoid skewing to anomalous or non-significant behaviours, all clusters should contain sufficient datapoints, and therefore the data sparsity of each cluster is checked. The sparsity of a cluster is defined as the ratio between the number of points it contains and the number of points in the largest cluster, $C_L$, leading to a dynamic sparsity measure that is relative to the current clustering. If the current largest cluster contains a relatively small number of datapoints from the entire dataset, the sparsity measure considers that all clusters are generally small and there may be intricate relationships within the data. The criteria for deciding whether cluster $C_k$ is sparse is shown in Equation \ref{eq:sparsity},
\begin{equation}
\label{eq:sparsity}
  \frac{|C_k|}{|C_L|} < \frac{1}{N^2}\sum_{i, j \in N} \lvert x_i - x_j \rvert,
\end{equation}
where N is the total number of samples in the dataset. The right-hand side of the formulation is the average pairwise distance between all points in the dataset and is used as a sparsity threshold since it provides a measure of the average density of the data. If the sparsity of a cluster falls below the threshold, it is combined with a neighbouring cluster in the feature space. The combination that gives the lowest clustering cost when merging the sparse cluster with each of its neighbours, is selected for the new clustering. Similarly, the coverage of each cluster is also checked, where coverage is calculated as the percentage of the input space occupied. If the coverage falls below the same threshold used for sparsity, it is combined with a neighbouring cluster using the same protocol as for a sparse cluster. 

To obtain a clustering, we start with some arbitrarily large $K$. An initial clustering is generated by selecting $K$ random medoids and assigning points to the cluster for which the medoid is most similar according to $\dissimilarity$. The cost of this clustering is calculated using Equation \ref{eq:clustering_cost}. A new clustering is generated by randomly selecting a new medoid for a random cluster and reassigning all points. If the new clustering is of lower cost, it replaces the previous clustering. This is repeated until the cost of the clustering is unchanged by selecting a new medoid. The clustering is then checked against the constraints and modified if necessary, which may lead to a change in the number of clusters. If so, $K$ is redefined as the new number of clusters and a new clustering is obtained in the same manner as outlined above, to find the lowest cost clustering for the new value of $K$. This process is repeated until the number of clusters does not change after satisfying the constraints. This algorithm is outlined in Appendix \ref{app:alg}.

\subsection{Generating Explanations}
We cluster each feature dimension in the input space and, as discussed in Section \ref{sec:problem_definition}, use this as a foundation for generating explanations for any input instance. The computational cost of MASALA scales linearly with the number of feature dimensions. A specified target instance $\instance$ will fall into a single linear region in each feature dimension. The set of instances $\localdata$ for training the surrogate $\explanation$ is defined using Equation \ref{eq:localdata}. The linear coefficients of the MLR indicate the contribution of each feature towards the base model's prediction for the target instance. 

\section{Experiments}
To evaluate MASALA we use the following two combinations of datasets and base models.

\textbf{PHM08 Challenge} \cite{saxenaDamagePropagationModeling} is a dataset used to predict the Remaining Useful Life (RUL) of a set of turbofan engines. A Gradient Boosting Regressor (GBR) \cite{friedmanElementsStatisticalLearning} was trained using the lifetime operations of 218 engines, containing almost 46,000 samples. GBRs provide global feature importance scores, however these describe general model behaviour and are not sufficient for providing insights into the behaviour of the model at an instance level. The GBR achieved a RMSE of 59.5 on the test set. Appendix \ref{app:PHM08_preds} compares the predictions made by the base model for the PHM08 dataset to the true values. It can be noticed that the relationship between features and predictions is not always linear, and therefore a single linear surrogate model may not be able to capture the true model behaviour.

\textbf{MIDAS} \cite{metoffice2019} is a dataset provided by the UK Meteorological Office which contains hourly weather observations across the UK. We use data collected at Heathrow Airport over 3 years (Jan. 2019 - Dec. 2022), which contains 19138 observations of a number of weather related parameters. A Recurrent Neural Network (RNN) is trained to predict the air temperature at a given time. RNNs are complex deep learning models that lack inherent-interpretability. Many state-of-the-art temporal models incorporate RNN architectures, therefore, being able to explain the behaviour of such models is useful to ensure they can be trusted and applied safely. The RNN achieved a RMSE of 3.04 on the test set. Appendix \ref{app:MIDAS_preds} compares the predictions made by the base model for the MIDAS dataset to the true values. Similar to the GBR, there is not a direct linear relationship between the distributions of input features and predictions.
For both datasets, 75\% of the data is used for training and 25\% is reserved for testing.

Local fidelity is a common evaluation metric that measures how well a surrogate model approximates the behaviour of the base model, by comparing their respective predictions on the local data used to fit the surrogate \cite{guidottiSurveyMethodsExplaining2019,freitasComprehensibleClassificationModels2014a}. This local data, as is the case for LIME and CHILLI, may be perturbed samples of the instance being explained \cite{anwarCHILLIDataContextaware2023,guidottiLocalRuleBasedExplanations2018}, which may not be appropriate if the perturbations are not representative of the original training data \cite{molnarGeneralPitfallsModelAgnostic2021}. Furthermore, locality is ill-defined in these existing methods and so local fidelity cannot be trusted. Instead, we measure {\em explanation fidelity} which is calculated as the absolute error between the predictions from the base model and surrogate model for the target instance.
\begin{equation}
\label{eq:explanation_fidelity}
    \text{Average Explanation Fidelity} = \frac{1}{N} \sum_{i=1}^{N} |f(\mathbf{x}_i) - \explanation(\mathbf{x}_i)|
\end{equation}

The average explanation fidelity can be calculated over a number of N instances, as shown in Equation \ref{eq:explanation_fidelity}, to quantify the explanation methods performance. We calculate the average explanation fidelity for 20 instances selectedly uniformly at random from the test set.

Prior works have highlighted that when repeatedly explaining the same instance, random perturbation based methods, such as LIME, produce inconsistent and differing explanations \cite{zafarDLIMEDeterministicLocal2019,zhangWhyShouldYou2019,liptonMythosModelInterpretability2017}. To measure consistency of repeated explanations, existing works use Jaccard distance \cite{zafarDLIMEDeterministicLocal2019,amparoreTrustNotTrust2021a}. Jaccard distance only considers explanations to be similar if their feature importance scores are identical. We instead propose calculating the average standard deviation of the normalised feature importance scores across 10 repeated explanations, which we subtract from 1 to measure {\em consistency}. 
We compare explanations generated by MASALA to those generated by LIME and CHILLI for a range of kernel width settings. Both the average consistency and fidelity of the explanations are calculated for all methods across 5 independent runs.

For illustration, the clustering used to generate explanations with MASALA for the MIDAS dataset is shown in Appendix \ref{app:midas_clustering}.
\section{Results}
\begin{figure}[t!]
\centering
\includegraphics[width=9cm]{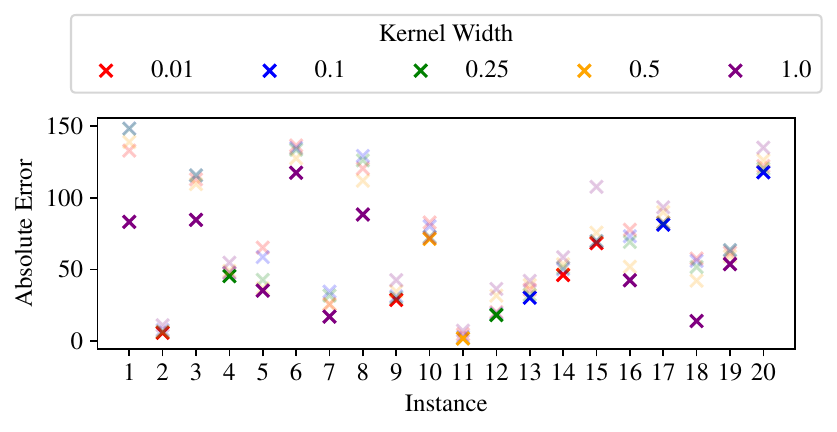}
\caption{\label{fig:error_per_instance} Error of explanations generated using CHILLI with different kernel widths for random instances from PHM08.}
\end{figure}

\begin{table}[ht!]
  \centering
  \begin{tabular}{@{\extracolsep{-3pt}}|l|c|c|c|c|}
      \hline
      \textbf{} & \multicolumn{2}{c|}{\textbf{Consistency}} & \multicolumn{2}{c|}{\textbf{Fidelity}} \\
      \textbf{} & \textbf{PHM08} & \textbf{MIDAS} & \textbf{PHM08} & \textbf{MIDAS}\\
      \hline
      \textbf{LIME} (0.1)     & 0.84 $\pm$ 0.134      & 0.54 $\pm$ 0.037      & 70.84 $\pm$ 15.199    & \textbf{0.0 $\pm$ 0.0} \\
      \textbf{LIME} (0.25)    & 0.63 $\pm$ 0.017      & 0.56 $\pm$ 0.029      & 70.84 $\pm$ 15.199    & 0.01 $\pm$ 0.004 \\
      \textbf{LIME} (0.5)     & 0.63 $\pm$ 0.02       & 0.84 $\pm$ 0.06       & 70.84 $\pm$ 15.199    & 1.16 $\pm$ 0.146 \\
      \textbf{LIME} (1.0)     & 0.73 $\pm$ 0.029      & 0.94 $\pm$ 0.055      & 66.95 $\pm$ 14.408    & 4.66 $\pm$ 0.39 \\
      \hline
      \textbf{CHILLI} (0.01)  & 0.97 $\pm$ 0.012      & 0.90 $\pm$ 0.070      & 67.21 $\pm$ 12.66     & \textbf{0.0 $\pm$ 0.0} \\
      \textbf{CHILLI} (0.1)   & 0.96 $\pm$ 0.014      & 0.97 $\pm$ 0.011      & 66.28 $\pm$ 12.224    & 0.75 $\pm$ 0.092 \\
      \textbf{CHILLI} (0.25)  & 0.95 $\pm$ 0.036      & 0.97 $\pm$ 0.009      & 65.36 $\pm$ 12.481    & 0.97 $\pm$ 0.15 \\
      \textbf{CHILLI} (0.5)   & 0.92 $\pm$ 0.013      & 0.97 $\pm$ 0.008      & 62.60 $\pm$ 11.163    & 1.03 $\pm$ 0.177 \\
      \textbf{CHILLI} (1.0)   & 0.89 $\pm$ 0.01       & 0.97 $\pm$ 0.009      & 60.53 $\pm$ 9.573     & 1.09 $\pm$ 0.184 \\
      \hline
      \textbf{MASALA}   & \textbf{1.00 $\pm$ 0.000}         & \textbf{1.00 $\pm$ 0.000}         & \textbf{35.94 $\pm$ 5.670}      & 0.50 $\pm$ 0.163 \\
      \hline
  \end{tabular}
  \caption{\label{tab:results}Average explanation consistency and fidelity achieved by LIME, CHILLI and MASALA with standard deviation across 5 independent runs.}
\end{table}
Table \ref{tab:results} shows the average consistency and fidelity of all explanations obtained from the experiments along with the standard deviation across the 5 runs. It can be noticed that on average, higher kernel width settings for LIME and CHILLI (shown in parentheses in Table~\ref{tab:results}) produce explanations with lower error for target instances from the PHM08 dataset. However, as the explanation fidelity improves, the consistency decreases which makes it difficult to trust the more performant explanations. Figure \ref{fig:error_per_instance} shows the absolute error of the explanations generated using CHILLI for the individual random instances at each kernel width setting. It can be noticed that the kernel width setting that produces the lowest error explanation varies. This highlights that the optimal locality parameter for each instance may vary and cannot be universally defined. For the PHM08 instances, explanations generated using MASALA always exhibited the lowest average error, and therefore greatest fidelity, compared to those generated using LIME and CHILLI across all kernel width values. This demonstrates the capability of MASALA to produce explanations that exceed the performance of LIME and CHILLI without the requirement of defining a kernel width value. Since the locality size of explanations generated using MASALA is dependent on local trends, it is able to capture the appropriate locality for individual instances without the need for a kernel width setting. 

\begin{figure*}
\includegraphics[width=15cm]{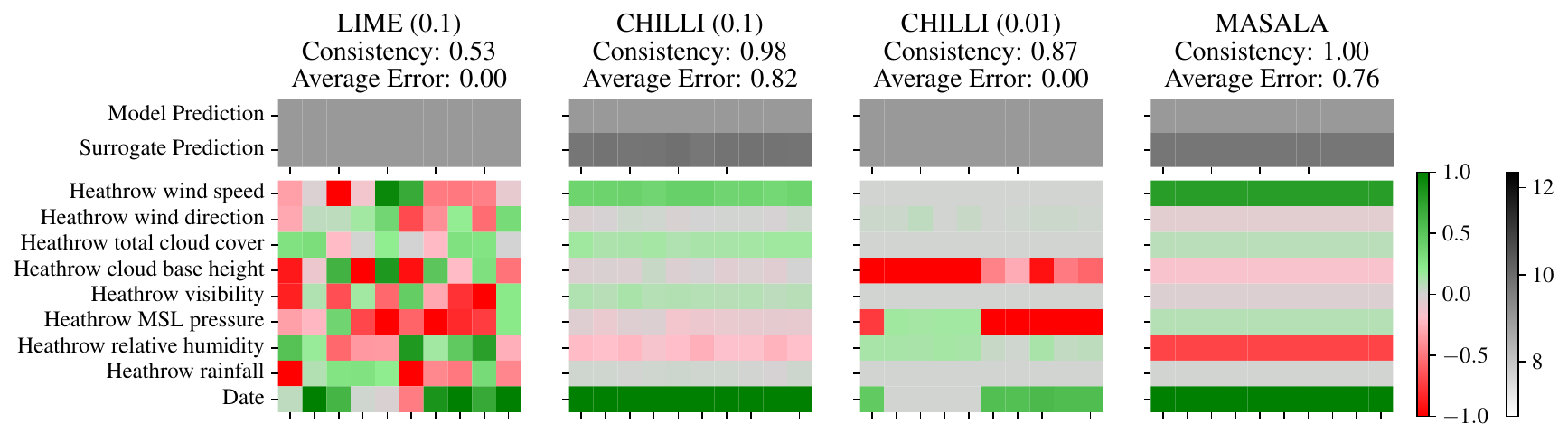}
\caption{\label{fig:example_explanations} Explanations generated using LIME, CHILLI and MASALA, 10 times for the same instance from the MIDAS dataset. Each column is a single explanation with the colour of the square indicating the linear relationship each feature has towards the target variable. The kernel width setting used for LIME and CHILLI is shown in parentheses.}
\end{figure*}

For the MIDAS dataset, CHILLI and LIME at low kernel width settings achieved significantly lower average error compared to MASALA. This can be attributed to the fact that at such small localities, the perturbations used to train the surrogate model only occupy a minuscule region of the input space, which the surrogate can model very accurately. It can also be noticed however, that at particularly small localities, LIME and CHILLI experience a lower consistency than MASALA. An example of repeated explanations for a single instance generated by each method is shown in Figure \ref{fig:example_explanations}. There is significant variation in explanations generated at low kernel width values which may be attributed to LIME and CHILLI randomly sampling perturbations. We see that for a kernel width of 0.1, CHILLI produces much more consistent explanations, since the perturbation generation method considers the distribution of the original data and this locality size may be appropriate in capturing the relevant trends. However, this comes at the cost of fidelity, indicated by the lower average error, since the produced explanations may be describing more general trends rather than local ones. Consistency cannot be sacrificed for fidelity, since if multiple explanations are presented for a single prediction, there is uncertainty regarding their correctness and trustworthiness. Explanations generated using MASALA achieve perfect consistency since they are generated using a predetermined clustering, leading to identical repeated explanations.

\section{Conclusion}
In this paper we propose MASALA, a novel method for generating explanations for black-box model predictions using linear local surrogate models.  We proposed a clustering technique that identifies a set of points which are similar to an instance for which an explanation is being generated, and use this to fit a linear surrogate model to approximate the base model behaviour. As a result, MASALA automatically detects the relevant and impactful model behaviour in an appropriately sized region of the input space. We find that explanations generated using our method produce more faithful and consistent explanations than those generated using LIME and CHILLI, without the need to manually define a locality hyperparameter that may differ for each instance being explained. Although a deterministic clustering ensures consistency, it is not clear how a non-deterministic clustering method which generates explaantions of equal fidelity would compare. There may be explanations that are equally faithful yet present different feature contributions so the question arises as to which of the explanations, or both, are correct. Future work would explore the possibilities of this and investigate whether such explanations are equally valid and how they can be compared.

\begin{acknowledgments}
We gratefully acknowledge the funding provided by the UK Engineering and Physical Sciences Research Council (grant ref. EP/T517641/1) and TRL Ltd to support this iCASE with project code B.CSAA.0001.
\end{acknowledgments}

\bibliography{journal_paper_refs}

\newpage
\appendix
\onecolumn
\section{Local Linear K-Medoids Clustering Algorithm}
\label{app:alg}
\begin{algorithm}[!h]
\caption{Local Linear K-Medoids Clustering}\label{alg:MASALA}
\begin{algorithmic}
\Require $K > 0$, $\dissimilarity$, Previous $K = 0$, Clustering Cost = $\infty$
\While{Previous $K \neq K$}
\State Previous $K = K$
\State Select $K$ medoids which are evenly distributed across the distribution of $\mathcal{X}$ 
  \For{each non-medoid $x \in \mathcal{X}$}
  \State Find the least dissimilar medoid according to $\dissimilarity$ and assign $x$ to 
  \State the corresponding cluster to generate clustering $C$
  \EndFor
  \State Fit LR model within each cluster $C_k$
\Repeat
  \State Calculate Clustering Cost $J(C)$, using Equation \ref{eq:clustering_cost}
  \State Lowest Cost = $J(C)$
  \For{ each cluster $C_k \in C$ }
  \For{each $x \in C_k$}
    \State Change medoid for $C_k$ to $x$
    \State Generate new clustering $C$ with new medoids
    \State Fit LR models within each cluster $C_k$
    \State Calculate $J'(C)$ using Equation \ref{eq:clustering_cost}
    \If {$J'(C) <$ Lowest Cost}
        \State Lowest Cost = $J(C)$
        \State Accept $x$ as new medoid
    \Else
        \State Reject $x$ as new medoid
    \EndIf
  \EndFor
  \EndFor
    \State Cost Difference = Lowest Cost - $J(C)$
    \If{Cost Difference < 0}
        \State Accept new clustering
        \State Clustering Cost = $J(C)$
    \EndIf
\Until Cost Difference = 0
\State $C'$ = Clusters after satisfying all constraints for clustering $C$
\State $K$ = Number of clusters in $C'$
\EndWhile
\end{algorithmic}
\end{algorithm}
\newpage
\section{MIDAS Clustering}
\label{app:midas_clustering}
Here we show the clustering obtained by applying Algorithm \ref{alg:MASALA} to the MIDAS data containing features describing weather characteristics at Heathrow Airport between 2019 and 2022. The distribution of each feature is shown against the predictions obtained by the base model being explained, in this case a Recurrent Neural Network. The clustered regions of linear behaviour are shown as different colours with the linear regression model fit to the points within each cluster also shown as a line of the same colour. 
\\
\\
We have not shown the clustering obtained for the PHM08 dataset because there are too many features to be able to visualise the individual clusters effectively.
\begin{figure}[!h]
\centering
  \includegraphics[width=15cm]{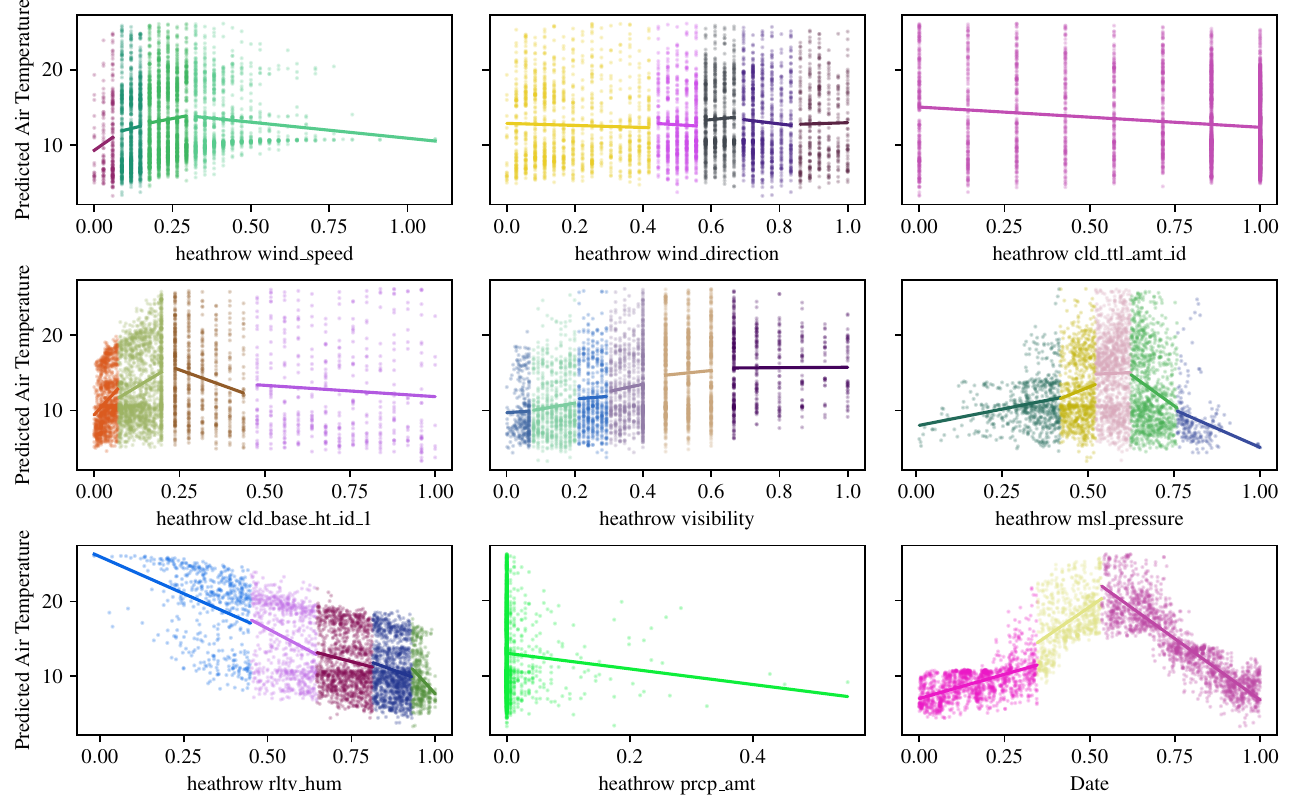}
\end{figure}

\newpage
\section{Model Predictions}

The distribution of each feature from the MIDAS and PHM08 datasets against the true target values and the RNN and GBR models respectively. It can be seen that there is not a clear linear relationship present between many features and the target variable. Therefore, a single linear model would not be appropriate for explaining the behaviour of the base model for all instances.

\subsection{MIDAS Recurrent Neural Network}
\label{app:MIDAS_preds}
\begin{figure}[!h]
\centering
  \includegraphics[width=15cm]{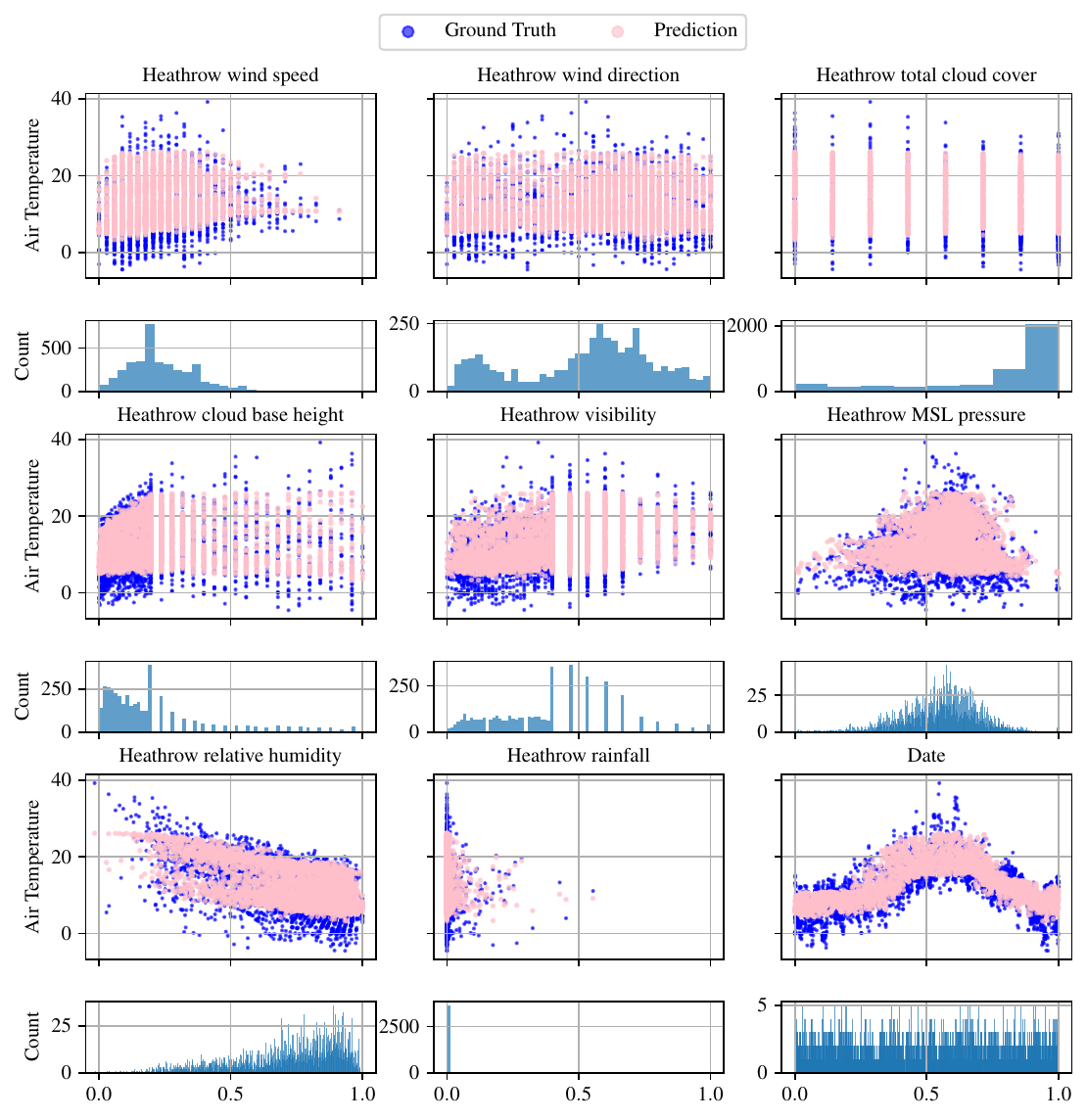}
\end{figure}

\newpage
\subsection{PHM08 Gradient Boosting Regressor}
\label{app:PHM08_preds}
\begin{figure}[!h]
\centering
  \includegraphics[width=13.5cm]{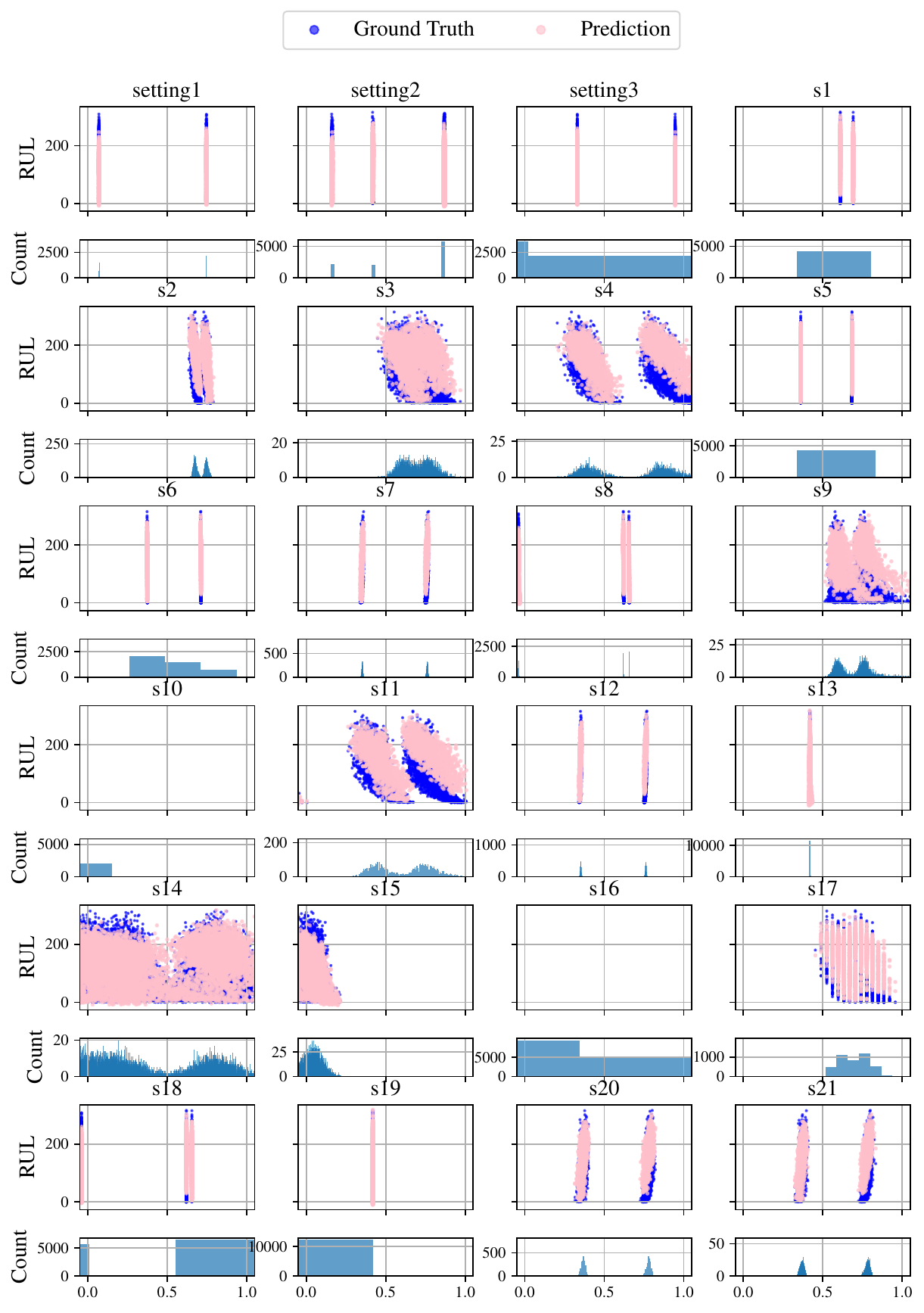}
\end{figure}

\end{document}